\documentclass[conference]{IEEEtran}
\IEEEoverridecommandlockouts
\usepackage{cite}
\usepackage{amsmath,amssymb,amsfonts}
\usepackage{algorithmic}
\usepackage{graphicx}
\usepackage{textcomp}
\usepackage{xcolor}
\usepackage{url}
\def\BibTeX{{\rm B\kern-.05em{\sc i\kern-.025em b}\kern-.08em
    T\kern-.1667em\lower.7ex\hbox{E}\kern-.125emX}}
\begin{document}

\title{Deploying Machine Learning Models to Ahead-of-Time Runtime on Edge Using MicroTVM
\thanks{
This work is funded by the German Research Foundation (DFG, Deutsche Forschungsgemeinschaft) as part of Germany’s Excellence Strategy – EXC 2050/1 – Project ID 390696704 – Cluster of Excellence “Centre for Tactile Internet with Human-in-the-Loop” (CeTI) of Technische Universit\"at Dresden.
}
\thanks{This paper was accepted at the CODAI'22 workshop, co-hosted with Embedded Systems Week, after it underwent a double-blind peer review process.}
}
\author{\IEEEauthorblockN{
Chen Liu\IEEEauthorrefmark{1},
Matthias Jobst\IEEEauthorrefmark{1}\IEEEauthorrefmark{2},
Liyuan Guo\IEEEauthorrefmark{1},
Xinyue Shi\IEEEauthorrefmark{1},
Johannes Partzsch\IEEEauthorrefmark{1}\IEEEauthorrefmark{2},
Christian Mayr\IEEEauthorrefmark{1}\IEEEauthorrefmark{2}
}
\IEEEauthorblockA{\IEEEauthorrefmark{1}\textit{Chair of Highly-Parallel VLSI-Systems and Neuro-Microelectronics} \\
\textit{Technische Universit\"at Dresden}\\
Dresden, Germany \\
Email: chen.liu@tu-dresden.de
}
\IEEEauthorblockA{\IEEEauthorrefmark{2}\textit{Centre for Tactile Internet with Human-in-the-loop (CeTI)} \\
\textit{Technische Universit\"at Dresden}\\
Dresden, Germany
}
}

\maketitle

\begin{abstract}
In the past few years, more and more AI applications have been applied to edge devices. However, models trained by data scientists with machine learning frameworks, such as PyTorch or TensorFlow, can not be seamlessly executed on edge. In this paper, we develop an end-to-end code generator parsing a pre-trained model to C source libraries for the backend using MicroTVM, a machine learning compiler framework extension addressing inference on bare metal devices.
An analysis shows that specific compute-intensive operators can be easily offloaded to the dedicated accelerator with a Universal Modular Accelerator (UMA) interface, while others are processed in the CPU cores.
By using the automatically generated ahead-of-time C runtime, we conduct a hand gesture recognition experiment on an ARM Cortex M4F core. 
\end{abstract}


\section{Introduction}
Machine learning has been significantly advanced in the past decade. 
With a rising number of AI solutions in consumer and industrial products,
AI tends to move closer to the edge, where is nearby the source of raw data.
Different from the typical training environment in the data center, edge devices focus more on inference, but under rigorous restrictions in terms of processing time, memory and power consumption, which hamper model deployment.

To bring AI to the edge, massive human efforts, such as graph optimization and quantization are mainly required. This process is prone to cause errors, to omit unconsidered operators, and is hard to be replicated on other devices, as it relies on the varied APIs provided by different backend chip vendors. Particularly when dedicated accelerators are involved to speed up specific operations, which is observed more and more common, model deployment becomes more complicated.

Machine learning compiler frameworks alleviate this issue.
They don't attempt to adapt the architecture of the model, but bridge the gap between a pre-trained model and a backend device by parsing, optimizing, and compiling the model graph, together with parameters, to executable.
Some sizable enterprises are keen on developing their own compiler frameworks, such as TensorFlow Lite (TFLite) \cite{tflite}, PyTorch Glow \cite{pyglow}, TVM \cite{tvm} and OpenVINO \cite{openvino}. 
These frameworks all support multiple input model formats and parse the model graph to high-level intermediate representatives (IR), then lower them to low-level IRs, which target heterogeneous backends like CPU, GPU, FPGA, and machine learning ASICs. 
However, a dynamical interpretation of the model graph is regarded as an inefficient way, especially on bare-metal Internet of things (IoT) devices. Ahead-of-time (AOT) compilation, as the alternative, draws booming interest due to less latency and more concise dependency.  
TensorFlow Lite Micro (TFLM) \cite{tflm} is the framework addressing AOT compilation on embedded systems among the TensorFlow toolbox, but requires C\texttt{++}11 support and comprises limited operators. Accordingly, TVM provides MicroTVM for the same sake.

In this paper we propose a MicroTVM based code generator, which automates the C source generation from a pre-trained model for a bare-metal backend chip. In case the model contains unsupported operators, we present a new operator registration strategy. In addition, an approach to offload specific operator patterns onto on-chip accelerators via a Universal Modular Accelerator (UMA) interface is discussed. The value of our tool is proved by deploying a deep neural network (DNN) into a real-time demonstration with a small implementation time, although no operators are currently ported to the accelerator for boosting processing performance.

The remaining of the paper will introduce the work flow of the code generator, associated with the unsupported operator registration to ensure the complete input model is parsable, and the UMA integration for intensive computation offloading. At last, the work flow is evaluated through an application of real-time hand gesture recognition.

\section{Implementation}

We implement an end-to-end C code generator with python language, which mainly uses the MicroTVM interface to translate the model graph down to C source code, together with periphery functionalities such as the generation of the top main function and default compiling script, i.e. makefile.

TVM has a frontend module, which naturally accepts almost all common DNN model formats. Currently the implemented frontend interface is for TensorFlow, TFLite, and ONNX, as they are widespread within the deep learning community, but it does not mean the tool is bounded to these 3 formats.
An expansion to other formats requires only a few lines of code. 
We choose structured C source code as the output of the generator to maximize multifarious backend devices, especially bare-metal IoT devices. At the same time, it provides flexibility by allowing custom modification on the generated source and compiling code.
Ideally, putting the generated directory at the proper position of the user's project, the application toolchain should be able to compile seamlessly.

\subsection{Overview}

\begin{figure}[htbp]
  \centering
  \includegraphics[width=\linewidth]{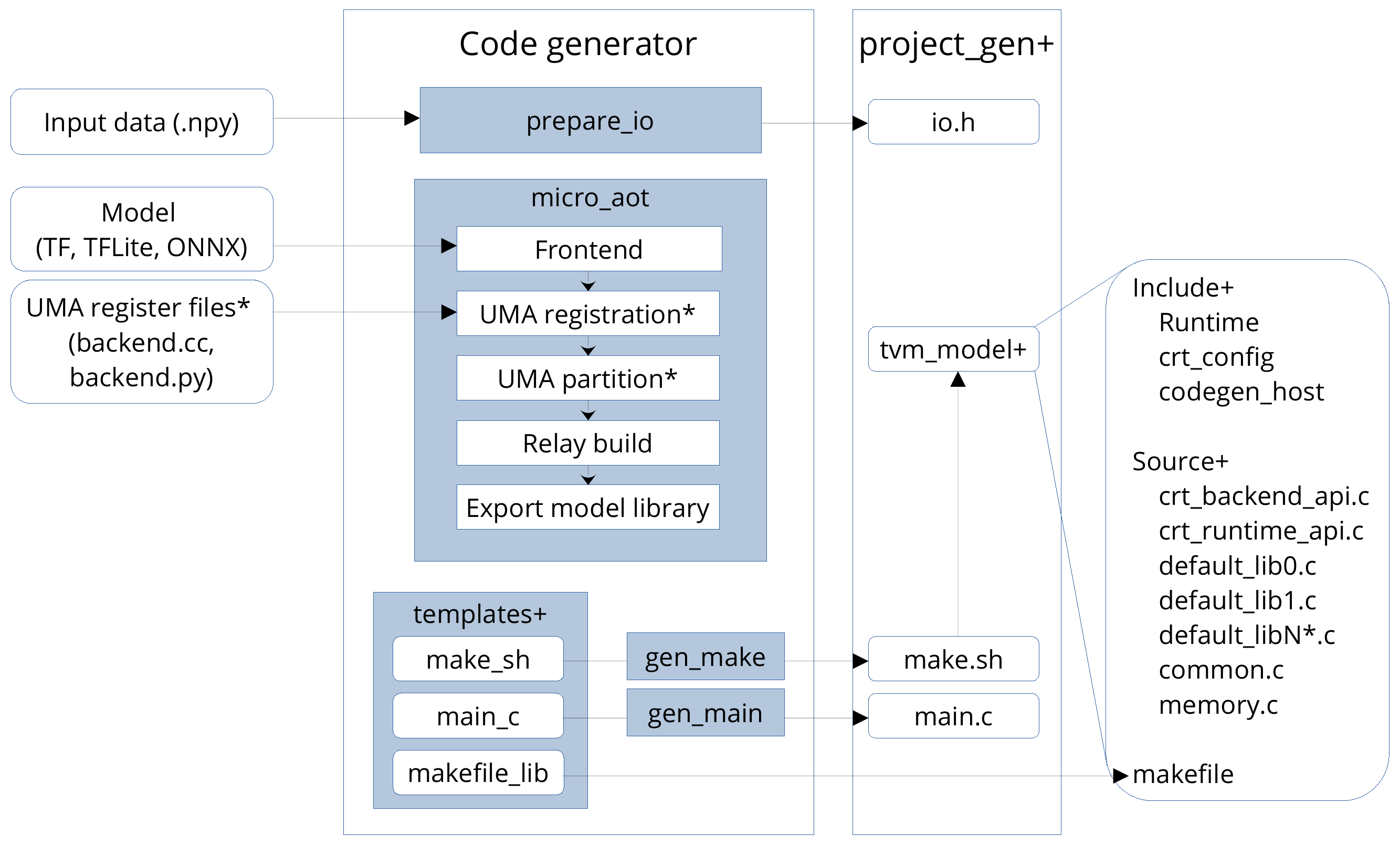}
  \caption{The architecture of the code generator. The sub-modules are marked with blue, files are wrapped in a rounded rectangle, and directories are suffixed with a plus sign.}
  \label{fig:codegen}
\end{figure}

The architecture of the code generator is illustrated in the block diagram of Fig.\ref{fig:codegen}. 
The {\itshape prepare\_io} module converts an input or output data sample file prepared by the user, to a C header file ({\itshape io.h}). 
The output data are taken as the reference during debugging of the inference result.
The {\itshape micro\_aot} module invokes MicroTVM in 5 phases. First, it uses the frontend interface to parse the prepared pre-trained model to a high-level Relay graph. If users intend to offload specific patterns of interest in the graph onto a dedicated computing node for acceleration, an accelerator needs to be registered via the UMA interface, which is a unified infrastructure for easily integrating external accelerators into TVM \cite{uma}. The graph is then correspondingly partitioned into the piece for the normal processor and the one for the accelerator. These 2 steps are optional and currently not completed, therefore they are marked with an asterisk sign and will be discussed in section \ref{sec:uma}. The Relay graph is lowered to the low-level tensor IR (TIR) through the build function in the Relay afterward. At last, those TIRs are transformed to C code in the form of plenty of header and source files. 

The contents of those template files in the {\itshape templates} directory are mainly static. During the generation, the {\itshape gen\_make} and {\itshape gen\_main} modules will duplicate the template files and replace preserved keywords with customized text for individualization. Since the Relay module exports not only the core source code but also a mass of metadata, collecting required files from distributed places is demanding. On the other hand, the application besides inference usually includes data preparation and communication with sensors or other devices. Assembling all key files and placing them in the {\itshape include} and {\itshape source} directories wraps the DNN as an isolated library, the {\itshape tvm\_model}. This process is executed in the generated {\itshape make.sh} file in the way of a sequence of commands. In addition, some compiling flags in the {\itshape makefile} of the library correlates strongly with the hardware, such as micro-architecture. Therefore its template ({\itshape makefile\_lib}) of varied backend should be accommodated in the {\itshape templates} directory as well.

In the usage, an input data file and a model file is the minimum requirement of the tool. The top executor of the code generator exports the primary transformed model, then the shell script further generates the organized {\itshape tvm\_module} and compiles the runtime binary with cross-compilation.

\subsection{New Operator Registration}
TVM is currently still under heavy development. It may occur that TVM cannot parse unknown operators contained in the pre-trained model, which can be not yet supported operators or custom operators, resulting in a failure at the frontend step. The frontend module for each specific format maintains a {\itshape Convert Map}, which is essentially a dictionary, whose key is the supported operator name, whereas the value is the conversion function interpreting the frontend operators with Relay tensor operators.

To register a new operator into the {\itshape Convert Map}, a new item with the operator name and corresponding conversion function should be appended. The latter can be implemented by calling existing Relay operators \cite{new_op3}, which is straightforward and doesn't require a recompiling of the TVM framework. Otherwise, a new relay operator should be created \cite{new_op1}\cite{new_op2}, which allows more flexibility but demands a recompiling of TVM.

\subsection{Accelerator Registration}\label{sec:uma}
\begin{figure}[htbp]
  \centering
  \includegraphics[width=\linewidth]{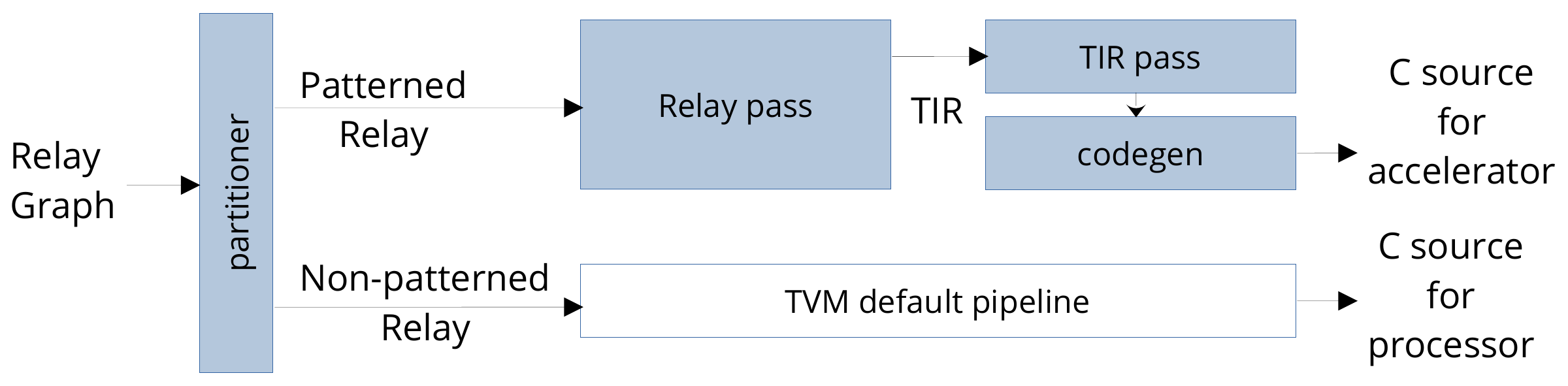}
  \caption{The UMA processing pipeline. Sub-procedures of UMA are marked with blue.}
  \label{fig:uma}
\end{figure}

Although TVM provides the Bring Your Own Codegen (BYOC) mechanism for hardware backend providers to implement their own code generator and register it as a Relay backend compiler to support dedicated accelerators, and there are already several accelerators that have been integrated into TVM, such as ARM Ethos NPU, incremental third-party advanced accelerators or new compute devices still being created, but not yet deeply involved in the TVM framework.

UMA as a consistent, unified infrastructure takes over the deep coupling work with TVM and provides python APIs to alleviate the accelerator registration \cite{uma}. As shown in Fig.\ref{fig:codegen}, the required input for UMA can be condensed into 2 files: A c/c++ file collecting the high-level and low-level C functions of the accelerator and a python file instantiating a UMA object by going through the UMA pipeline, which is exhibited in Fig.\ref{fig:uma}. 

Given a Relay graph is successfully extracted from the frontend, The partitioner of the UMA splits Relay operators into 2 groups according to the registered pattern. The non-patterned operators go through the default TVM pipeline as described before and translated as the runtime in the processor, while the patterned operators are processed with the Relay pass, like configurations or optimizations. After they are lowered down to TIR, the TIR pass handles for example parameters and data formats and converts them to the prime functions, which are finally interfaced with the registered C functions in the c/c++ file invoked by the backend's accelerator.

So far we leave related API in the code generator to integrate UMA for offloading compute-intensive operators in DNN, like two-dimensional convolution and matrix multiplication, but testing in the real accelerator is still ongoing.

\section{Evaluation}

We evaluate the code generator for an agile DNN model deployment without UMA on a bare-metal embedded chip. A neural network model is trained for a hand motion gesture classification task with 20 different motion gestures, based on 6D inertial measurement unit (IMU) data, utilizing acceleration and gyroscope data. For training, we use the 6D Motion Gesture Database (6DMG) \cite{6DMG} consisting of 5600 labeled gesture recordings from 28 persons.

\subsection{Evaluation Setup}
In the real-time setup, we use an experimental tactile glove mounted with an IMU on the back of the hand \cite{ceti_glove} to collect the hand motion information, which is then transmitted via Bluetooth to a laptop. The laptop forwards the data to the connected backend chip, in the meanwhile, runs a user interface (UI) program to showcase the classification result. 

\begin{figure}[htbp]
  \centering
  \includegraphics[width=\linewidth]{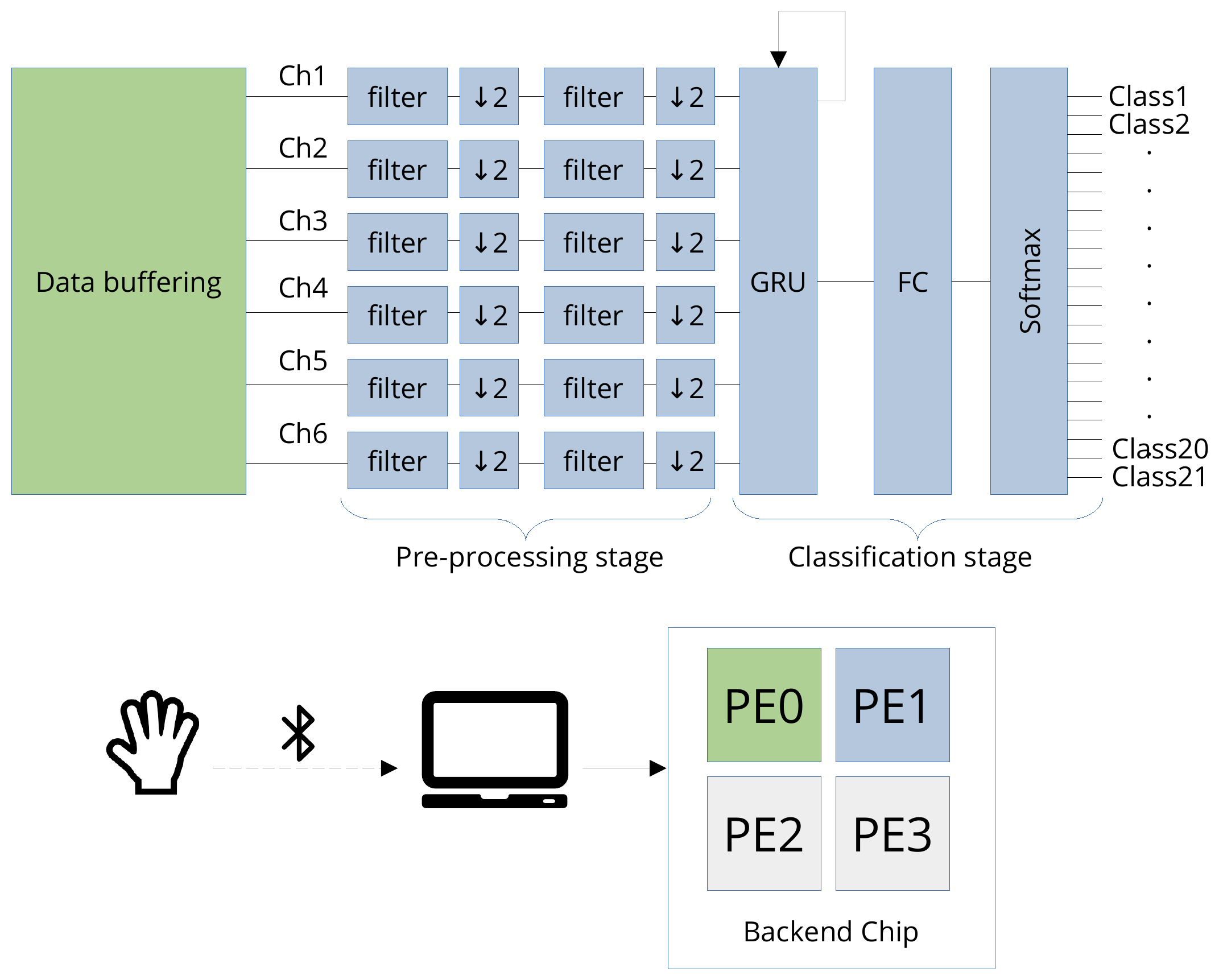}
  \caption{Data flow on the evaluation setup.}
  \label{fig:data flow}
\end{figure}

The adopted backend chip is a distributed multi-core platform \cite{jib1}. Each processing element (PE) consists of a Cortex-M4F processor allocated with 128~kB SRAM and a series of arithmetic accelerators for such as exponential and multiply-accumulate (MAC) operations.
Each PE can be flexibly activated depending on the application. We employ 2 PEs to classify the real-time hand motion gesture. The first PE buffers the received data from outside. Once data sufficient for one inference is collected, they are transferred to the specific memory section of the neighbor PE via direct memory access (DMA). The second PE loads the generated C runtime for classification. It will be triggered when DMA is done and return the classified output to the laptop for result visualization. The data flow on the evaluation setup is illustrated in Fig.\ref{fig:data flow}.

\subsection{Evaluation Model}
The input data of the model is a time series of 6 channels and varying lengths, where the mean length varies between 743.6 and 1806.5\,ms, depending on the gesture. In order to conduct the inference for real-time usage, we extend the dataset by a gesture class of "none". The data of this class is generated  by adding a small amount of noise on top of static offsets in each channel, where the offset is random for each generated recording.

A tiny recurrent neural network (RNN) very similar to \cite{zen_aicas} is employed. The two-stage approach consists of a feature extraction stage with cascaded channel-wise convolutional filters and a classification stage with recurrent and feed-forward units. 
Specifically, the filter bank is composed of two one-dimensional filter stages, with a filter length of 7 and a stride of 2 at each output, reducing the data rate for the RNN by a factor of 4. 
Each filter stage shares the parameters over all 6 channels. 
The classifier is a composition of gated recurrent units (GRU) with 16 units and a fully-connected (FC) layer with one softmax output neuron per gesture, resulting in 21 outputs.

The entire network has a total of only 1525 parameters, which can be accommodated in a single PE. After 100 epochs of training with the Adam optimizer \cite{adam} it achieves an accuracy of 96.6\,\% on a separate test dataset aside from the main dataset for training. 

Quantization is a commonly used technique to save model size and inference latency at the deployment. However, as a floating-point unit (FPU) is integrated in the backend chip and the model is so tiny that can be fully loaded into memory with single-precision floating point, quantization is not applied in the evaluation, and the generated C source does not depend on the CMSIS-NN library \cite{cmsis} accordingly.
\subsection{Evaluation Result}
In the experiment we separate the processing chain into data management (pre- and post-processing) and DNN inference. Combined with the characteristics of multiple computation nodes in the backend chip, the 2 processing segments are mapped to the management PE (PE0) and the classification PE (PE1), respectively. The C runtime on the classification PE is directly generated from our code generator, instead of the tedious manual model implementation. Only minor code modification is needed to make it compatible with the management runtime, when they are wrapped up to the top runtime.

\begin{figure}[htbp]
  \centering
  \includegraphics[width=\linewidth]{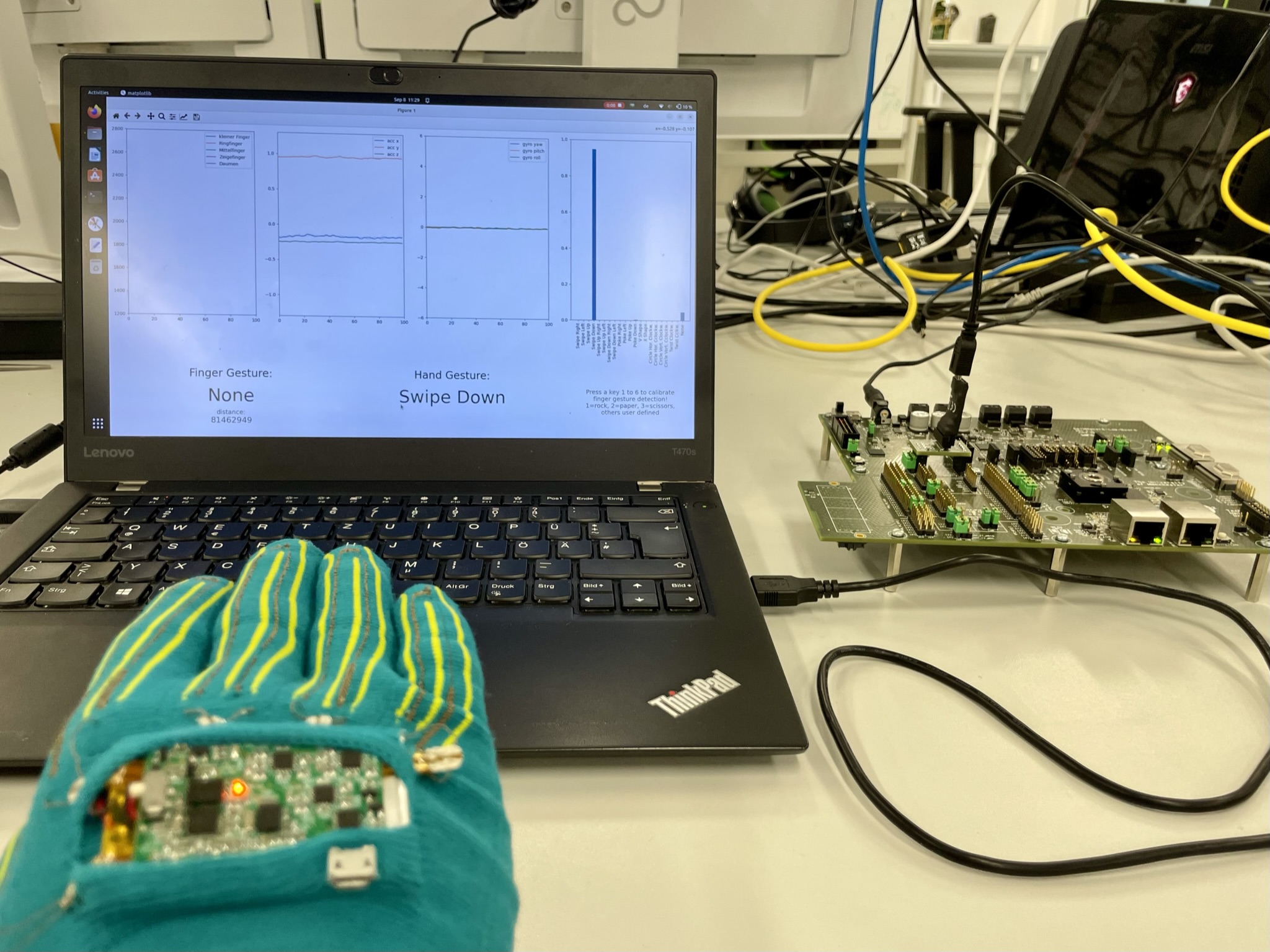}
  \caption{Real-time demo of hand gesture recognition.}
  \label{fig:demo}
\end{figure}

A live demonstration is shown in Fig.\ref{fig:demo}. The UI dashboard exhibits the finger gesture and hand gesture, where the latter is one of the 21 classes output from the model. It is observed when the subject is swiping his hand down, the classifier recognizes the action of "swipe down" with high confidence.

\section{Conclusion}
Based on TVM framework we build a code generator and wrap it as a python library, targetting an agile DNN model deployment on the edge from end to end. A demonstration of the real-time hand gesture recognition exhibits that our tool liberates much effort for DNN model implementation on the edge device, only requiring minor modifications for the work of data management. Our tool also leaves an interface with an external accelerator via UMA. The performance improvement of the same inference after the integration of an accelerator will be tested in the future.

\bibliographystyle{IEEEtran}
\bibliography{IEEEabrv,camera_ready.bib}

\end{document}